\documentclass{article}

\usepackage[numbers]{natbib}
\usepackage[preprint]{neurips_2025_custom}

\usepackage{amsmath,amsfonts,bm}

\def\eqref#1{equation~(\ref{#1})}
\def\Eqref#1{Equation~(\ref{#1})}

\def\1{\bm{1}}

\DeclareMathAlphabet{\mathsfit}{\encodingdefault}{\sfdefault}{m}{sl}
\SetMathAlphabet{\mathsfit}{bold}{\encodingdefault}{\sfdefault}{bx}{n}

\usepackage[dvipsnames]{xcolor}         
\definecolor{linkColor}{rgb}{0.18,0.39,0.62}
\usepackage[utf8]{inputenc} 
\usepackage[T1]{fontenc}    
\usepackage[colorlinks=true,linkcolor=linkColor,citecolor=linkColor,filecolor=linkColor,urlcolor=linkColor]{hyperref}       
\usepackage{cleveref}
\usepackage{multirow}
\usepackage{xspace}
\usepackage{booktabs}
\usepackage{graphicx}
\usepackage{array}
\usepackage{wrapfig}
\usepackage{bm}
\usepackage{algorithm}

\usepackage{algpseudocode}
\usepackage{algpseudocode}
\usepackage{enumitem}
\usepackage{tcolorbox}
\usepackage{makecell}
\usepackage{diagbox}

\usepackage{amsmath}
\usepackage{amssymb}
\usepackage{mathtools}
\usepackage{amsthm}

\usepackage{multirow}
\usepackage{amsmath}
\usepackage{capt-of}
\usepackage{tabularx}
\usepackage{epsfig}
\usepackage{amssymb}
\usepackage{amsfonts}
\usepackage{booktabs}
\usepackage{scalerel}
\usepackage{listings}
\usepackage{varwidth}
\usepackage{stmaryrd}
\usepackage{bbm}
\usepackage{wrapfig}
\usepackage{pifont}

\newcommand{\tabincell}[2]{\begin{tabular}{@{}#1@{}}#2\end{tabular}}

\definecolor{deepblue}{rgb}{0,0,0.5}
\definecolor{officeblue}{RGB}{0,102,204}
\definecolor{deepred}{rgb}{0.6,0,0}
\definecolor{deepgreen}{rgb}{0,0.5,0}
\definecolor{mybrickred}{RGB}{182,50,28}

\definecolor{fillcolor}{RGB}{216,217,252}

\usepackage{etoolbox}
\usepackage{framed}

\newif\ifxetexorluatex
\ifxetex
  \xetexorluatextrue
\else
  \ifluatex
    \xetexorluatextrue
  \else
    \xetexorluatexfalse
  \fi
\fi
\newcommand*\quotesize{60} 
\newcommand*{\openquote}
   {\tikz[remember picture,overlay,xshift=-4ex,yshift=-2.5ex]
   \node (OQ) {\fontsize{\quotesize}{\quotesize}\selectfont``};\kern0pt}

\newcommand*{\closequote}[1]
  {\tikz[remember picture,overlay,xshift=4ex,yshift={#1}]
   \node (CQ) {\fontsize{\quotesize}{\quotesize}\selectfont''};}

\colorlet{shadecolor}{white}

\newcommand*\shadedauthorformat{\emph} 

\newcommand*\authoralign[1]{%
  \if#1l
    \def\authorfill{}\def\quotefill{\hfill}
  \else
    \if#1r
      \def\authorfill{\hfill}\def\quotefill{}
    \else
      \if#1c
        \gdef\authorfill{\hfill}\def\quotefill{\hfill}
      \else\typeout{Invalid option}
      \fi
    \fi
  \fi}
{\authoralign{#1}
\ifblank{#2}
   {\def\shadequoteauthor{}\def\yshift{-2ex}\def\quotefill{\hfill}}
   {\def\shadequoteauthor{\par\authorfill\shadedauthorformat{#2}}\def\yshift{2ex}}
\begin{snugshade}\begin{quote}\openquote}
{\shadequoteauthor\quotefill\closequote{\yshift}\end{quote}\end{snugshade}}

\newcommand{\github}{\raisebox{-1.5pt}{\includegraphics[height=1.05em]{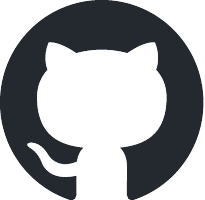}}\xspace}

\definecolor{DarkBlue}{RGB}{0, 51, 153}
\newcommand{\ours}{OEL}

\title{Online Experiential Learning for Language Models}

\author{%
Tianzhu Ye\thanks{~Equal contribution.}~~~~~~~~Li Dong\footnotemark[1] \\
\bf Qingxiu Dong~~~~~Xun Wu~~~~~Shaohan Huang~~~~~Furu Wei \\
~Microsoft Research \\
~{\href{https://aka.ms/GeneralAI}{https://aka.ms/GeneralAI}}
}

\begin{document}

\maketitle

\footnotetext[1]{This is Part~II of Experiential Learning series. Part~I: \href{https://arxiv.org/abs/2602.12275}{On-Policy Context Distillation for Language Models}.}

\begin{abstract}
The prevailing paradigm for improving large language models relies on offline training with human annotations or simulated environments, leaving the rich experience accumulated during real-world deployment entirely unexploited. We propose \textbf{Online Experiential Learning} (\ours{}), a framework that enables language models to continuously improve from their own deployment experience. \ours{} operates in two stages: first, transferable experiential knowledge is extracted and accumulated from interaction trajectories collected on the user side; second, this knowledge is consolidated into model parameters via on-policy context distillation, requiring no access to the user-side environment. The two stages are iterated to form an online learning loop, where the improved model collects higher-quality trajectories that yield richer experiential knowledge for subsequent rounds. We evaluate \ours{} on text-based game environments across multiple model scales and both thinking and non-thinking variants. \ours{} achieves consistent improvements over successive iterations, enhancing both \textbf{task accuracy} and \textbf{token efficiency} while preserving out-of-distribution performance. Our analysis further shows that extracted experiential knowledge is significantly more effective than raw trajectories, and that on-policy consistency between the knowledge source and the policy model is critical for effective learning.
\begin{table}[H]
\centering
\begin{tabular}{@{}r@{\hspace{2pt}}l@{}}
\github & \textbf{Code}: \href{https://aka.ms/oel-code}{\texttt{aka.ms/oel-code}}
\end{tabular}
\end{table}
\end{abstract}

\vfill{}

\begin{figure}[ht]
\centering
\includegraphics[width=0.99\linewidth]{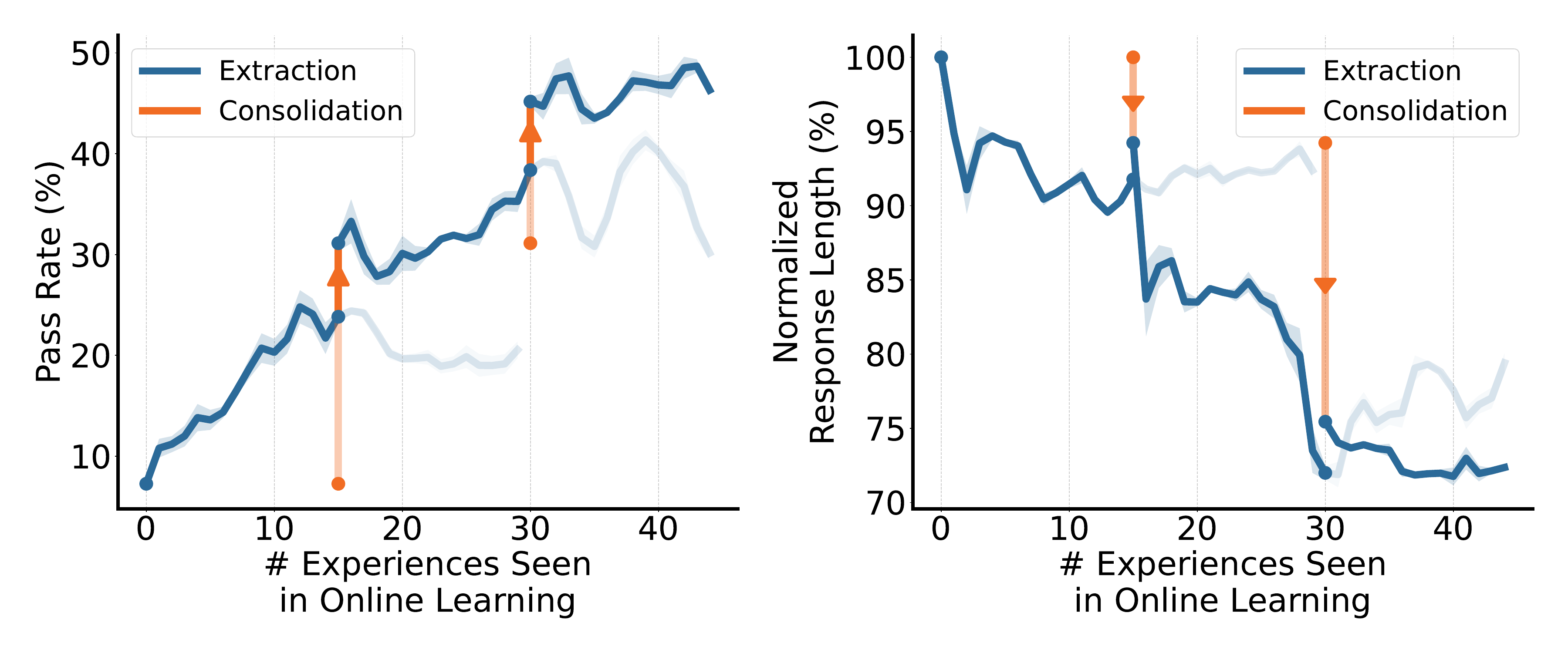}
\vspace{-0.1cm}
\caption{By iterating over experiential knowledge extraction and consolidation stages of \ours{}, the model can progressively improve pass rate and efficiency (measured by response length) on the environment, effectively achieving online learning.}
\vspace{-0.1cm}
\label{fig:oel_curve}
\end{figure}

\vfill{}

\newpage
\section{Introduction}
\label{sec:intro}
Large language models (LLMs) have demonstrated remarkable capabilities across a wide range of tasks, from mathematical reasoning to code generation and open-ended dialogue~\citep{gpt4,qwen3,deepseekr1}. Yet the dominant approach to improving these models remains fundamentally \emph{offline}: practitioners collect human annotations for supervised fine-tuning, or construct simulated environments with verifiable rewards for reinforcement learning~\citep{grpo,dapo}. The model is trained and deployed as a static artifact. While effective within its training distribution, the paradigm creates an inherent bottleneck---the model can only be as good as the data and environments curated before deployment. Once deployed, the model encounters a vast, ever-evolving landscape of real-world tasks and user needs, yet gains nothing from these interactions. The rich stream of experience accumulated during deployment is simply discarded.

We envision a paradigm of \textbf{online learning} where the model does not stop improving after deployment, but instead continues to learn from its interactions with real-world environments, progressively refining its capabilities over time. Yet realizing this vision is far from straightforward. The server side, where model training takes place, typically cannot access the user-side environments in which the model operates. Furthermore, real-world interactions rarely provide scalar reward signals; instead, the environment returns only textual feedback such as natural language descriptions of outcomes, errors, or state changes. Standard reinforcement learning algorithms cannot directly consume such unstructured signals, and constructing verifiable reward functions or training reward models for every new deployment scenario is impractical. These constraints demand a new learning paradigm that can extract useful training signal from raw textual experience alone, without requiring environment access or reward supervision on the server side.

In this work, we propose \textbf{Online Experiential Learning} (\ours{}), a framework that enables language models to continuously improve from their own deployment experience. The key insight is to convert textual environment feedback into \emph{experiential knowledge} that can be extracted, accumulated, and internalized into model parameters. \ours{} operates in two stages. In the first stage, the model extracts transferable experiential knowledge from interaction trajectories collected during deployment, accumulating insights across multiple episodes. In the second stage, this accumulated knowledge is consolidated into the model's parameters via on-policy context distillation \citep{opcd}, which trains the model to match the behavior of a knowledge-conditioned teacher without requiring the knowledge context at inference time. Crucially, the entire process is reward-free: no reward model, no verifiable reward function, and no human annotation is needed. On the user side, the only requirement is to collect interaction trajectories during normal usage; on the server side, training is carried out entirely from these pre-collected trajectories without access to the user-side environment. The two stages can be iterated: the improved model is redeployed to collect higher-quality trajectories, yielding richer experiential knowledge for the next round of consolidation, naturally forming an online learning loop.

We evaluate \ours{} on two environments. Across multiple model scales and both thinking and non-thinking model variants, \ours{} achieves consistent and substantial improvements over successive iterations. We further demonstrate that \ours{} improves not only task accuracy but also inference efficiency, with response lengths decreasing as experiential knowledge is internalized. Importantly, the on-policy context distillation used in \ours{} preserves out-of-distribution performance, mitigating catastrophic forgetting compared to off-policy alternatives. Our analysis reveals that extracted experiential knowledge is significantly more effective than raw trajectories, and that on-policy consistency between the knowledge source and the policy model is critical for effective learning.

\section{Preliminary: Online Learning}
\label{sec:owol}

\begin{figure}[t]
\centering
\includegraphics[width=0.95\linewidth]{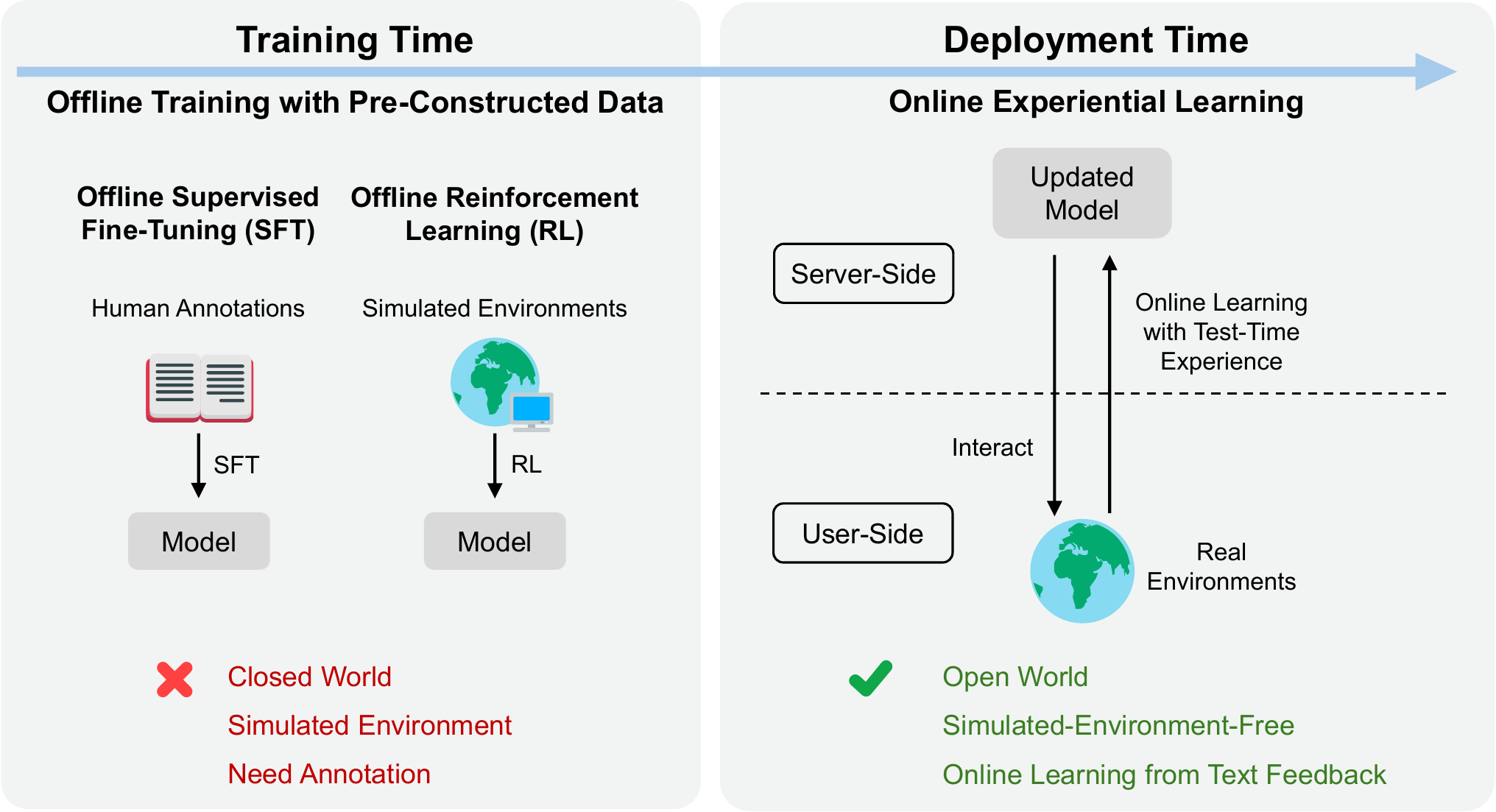}
\vspace{-0.1cm}
\caption{\textbf{Offline training vs.\ online experiential learning.} \textit{Left:} The prevailing offline paradigm trains models at the server side using human annotations (SFT) or simulated environments (RL), operating in a closed world with pre-constructed data. \textit{Right:} Online experiential learning forms a virtuous cycle during deployment. The model interacts with real environments on the user side, and the resulting test-time experience is used to update the model on the server side, requiring no annotations, no simulated environments, and enabling open-world learning from text feedback.}
\vspace{-0.1cm}
\label{fig:ol}
\end{figure}

As large language models are increasingly deployed across diverse real-world scenarios, they inevitably encounter an open-ended stream of environments, tasks, and user demands that far exceed what any controlled training setting can anticipate. As illustrated in \Cref{fig:ol} (left), the prevailing paradigm relies on \emph{offline} training with pre-constructed data: supervised fine-tuning with human annotations and reinforcement learning with verifiable rewards or reward models in simulated environments. While effective for targeted optimization, this offline paradigm faces a fundamental ceiling---performance saturates on the curated training distribution, and further scaling requires increasingly costly annotations or increasingly faithful simulations, neither of which can fully cover the diversity of real-world deployment.

We advocate for \textbf{online experiential learning} as a fundamentally scalable paradigm (\Cref{fig:ol}, right). Rather than relying on offline-constructed supervision, this paradigm leverages the test-time experience that the model naturally accumulates through interactions with real environments as the primary signal for improvement. Crucially, this approach is \emph{reward-free}: it requires no human annotations, no verifiable reward functions, and no simulated environments on the server side. The model is deployed and interacts with users in the open world; the resulting experience is then fed back to update the model. Deployment and learning are thus connected in a virtuous cycle---the broader the deployment, the richer the signal for continued improvement. We believe this paradigm will become essential for the next stage of LLM development, as real-world deployment offers a virtually unlimited and ever-evolving source of learning signal that offline training alone cannot substitute.

\section{Online Experiential Learning}
\label{sec:method}

\begin{figure}[t]
\centering
\includegraphics[width=0.95\linewidth]{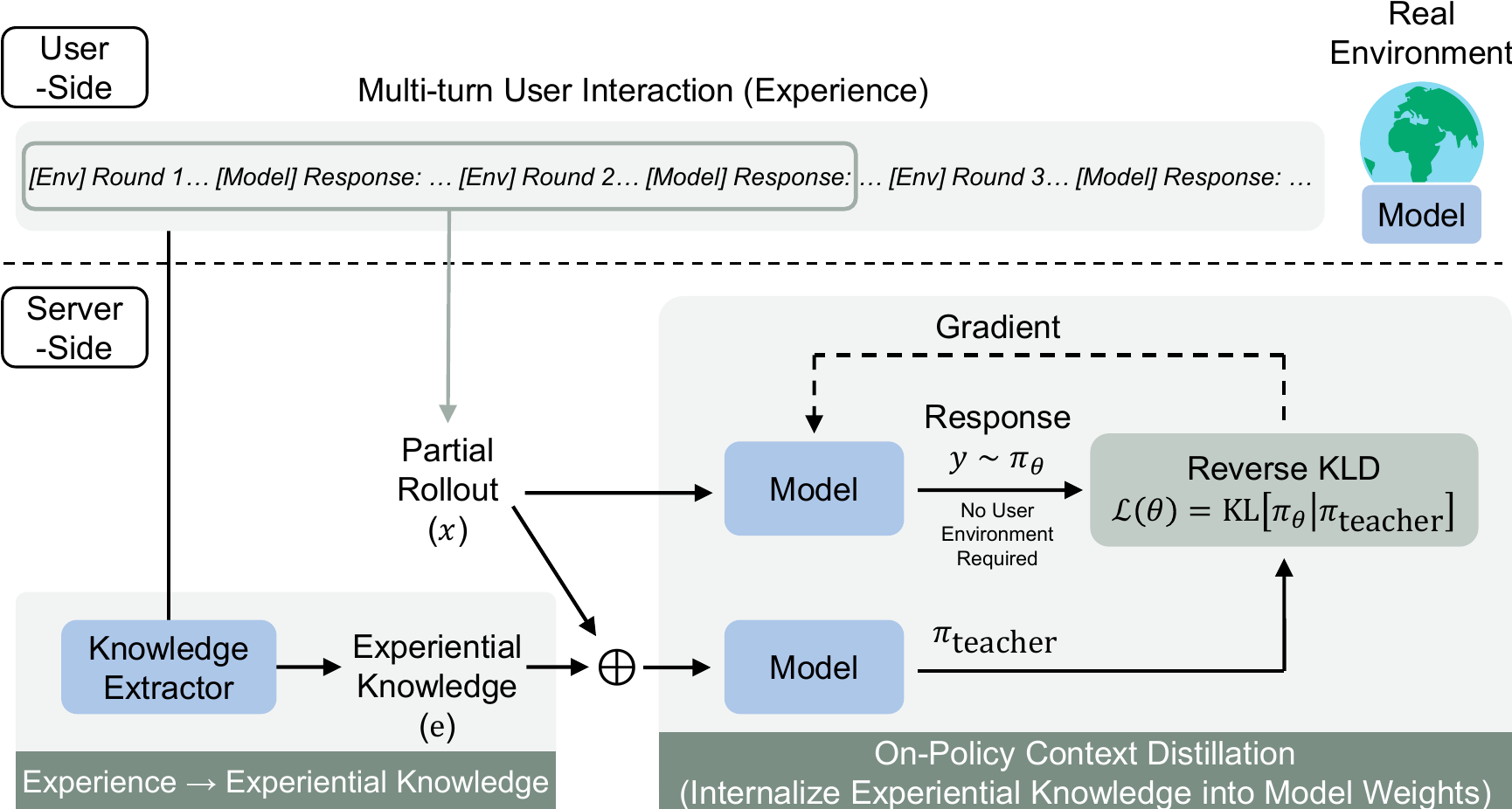}
\caption{\textbf{Overview of \ours{}.} On the user side, the model interacts with the real environment to collect multi-turn trajectories. On the server side, transferable experiential knowledge is first extracted from the collected trajectories, then consolidated into model weights via on-policy context distillation. During training, the model performs single-turn rollouts from partial rollout prefixes and is optimized to match a knowledge-conditioned teacher through reverse KL divergence, eliminating the need for user-side environment access. The entire process relies solely on textual environment feedback, requiring no reward model or verifiable reward.}
\label{fig:method}
\end{figure}

We present \textbf{Online Experiential Learning} (\ours{}), a framework illustrated in \Cref{fig:method}.
On the user side, the model interacts with the real environment to collect multi-turn trajectories.
Then on the server side, the learning proceeds in two stages: first, transferable experiential knowledge is extracted from the collected trajectories; second, this knowledge is consolidated into the model parameters via on-policy context distillation~\citep{opcd}, where the model generates single-turn responses from partial rollouts and is trained to match a knowledge-conditioned teacher through reverse KL divergence—without requiring access to the user-side environment.

Notably, \ours{} enables on-policy learning using only textual environment feedback, requiring no reward model or verifiable reward. As the model improves, it collects higher-quality trajectories that yield richer experiential knowledge, which in turn drives further improvement. This process can be iterated to progressively improve performance, forming an online learning loop (\Cref{sec:iteration}).

\subsection{Extract Experiential Knowledge from User Trajectories}

We consider a language model $\pi_\theta$ deployed to interact with a user-side environment $\mathcal{E}$. It collects a set of $n$ trajectories, $\mathcal{T} = \{\tau_1, \tau_2, \dots, \tau_n\}$, 
where each trajectory $\tau_i = (f_i^1, a_i^1, f_i^2, a_i^2, \ldots)$ consists of an alternating sequence of model actions and textual environment feedback.
Given the collected trajectories, we employ a language model $\pi_\mathrm{extract}$ to sequentially extract transferable experiential knowledge learned from each trajectory. By default we use $\pi_\mathrm{extract}=\pi_\theta$. The extraction proceeds in an accumulative fashion: when processing the $i$-th trajectory, the model also conditions on previously accumulated experiential knowledge.

Formally, let $e_i$ denote the accumulated experiential knowledge after processing trajectory $\tau_i$, with $e_0 = \emptyset$. The extraction and accumulation process is defined recursively for $i = 1, \dots, n$ as:

\begin{equation}
\begin{gathered}
e_i' \sim \pi_{\mathrm{extract}}(\cdot \mid \tau_i,\, e_{i-1}) \\
e_i = [e_{i-1};\, e_i']
\label{eq:extraction}
\end{gathered}
\end{equation}
where $[e_{i-1};\, e_i']$ denotes the concatenation of the previous accumulated experiential knowledge and the newly extracted knowledge from $\tau_i$.
Notably, this extraction process does not rely on ground-truth labels; the model conditions solely on interaction trajectories with the user-side environment.

\subsection{Consolidate Experiential Knowledge into Model Weights}

After extraction, we obtain a set of experiential knowledge $\mathcal{C} = \{e^1, e^2, \ldots, e^K\}$, where each $e^k$ is produced by running the accumulation process over $\mathcal{T}$ with a different random seed. We then consolidate this knowledge into the model parameters via on-policy context distillation~\citep{opcd}.

Specifically, the user collects $m$ interaction trajectories $\mathcal{T}' = \{\tau_1, \tau_2, \ldots, \tau_m\}$ from the environment $\mathcal{E}$. From each trajectory $\tau_i$, we extract all partial rollout prefixes $x_i^j = (f_i^1, a_i^1, \ldots, f_i^{j-1}, a_i^{j-1}, f_i^j)$, each capturing the interaction history up to but not including the $j$-th model response. The full set of prefixes across all trajectories forms the training dataset $\mathcal{D} = \{x_i^j\}$. During training, the model performs a single-turn response generation conditioned on each prefix, which enables on-policy learning without requiring access to the user-side environment.

On the server side, we train the model $\pi_\theta$ to internalize the experiential knowledge via on-policy context distillation. For each training step, we sample prefix $x$ from $\mathcal{D}$ and experiential knowledge $e$ from $\mathcal{C}$. The student $\pi_\theta$ generates a response $y$ conditioned only on $x$, and is optimized to match the knowledge-conditioned output of a teacher $\pi_{\mathrm{teacher}}$ through token-level reverse KL divergence~\citep{minillm}:
\begin{equation}
\mathcal{L}(\theta) = \mathbb{E}_{x \sim \mathcal{D}, e \sim \mathcal{C}, y \sim \pi_\theta(\cdot \mid x)} \left[ \frac{1}{|y|} \sum_{t=1}^{|y|} 
D_{\mathrm{KL}}\!\left( \pi_\theta(\cdot \mid x, y_{<t}) \,\Big\|\, 
\pi_{\mathrm{teacher}}(\cdot \mid e, x, y_{<t}) \right) \right]
\label{eq:objective}
\end{equation}
We use the frozen initial $\pi_\theta$ before training as $\pi_{\mathrm{teacher}}$ in this work. Since the model performs single-turn rollouts at each response position, \textbf{the entire training procedure can be carried out on the server side without access to the user-side environment $\mathcal{E}$}. Moreover, the experiential-knowledge-conditioned teacher provides dense, token-level training signal derived solely from \textbf{textual environment feedback collected on the user side, requiring no reward model or verifiable reward}. Refer to Appendix~\ref{app:opcd_detail} for more details.

\subsection{Online Learning Process}
\label{sec:iteration}


The two stages described above can be naturally iterated to progressively improve model performance.
After each round of consolidation, the updated model $\pi_\theta$ is deployed back to the user-side environment $\mathcal{E}$ to collect a new set of trajectories $\mathcal{T}$ and $\mathcal{T'}$.
As the model improves, the newly collected trajectories reflect higher-quality behavior, yielding richer experiential knowledge upon extraction.
This accumulated knowledge $\mathcal{C}$ is then used to drive the next round of consolidation, creating a virtuous cycle where better models produce better trajectories, which in turn yield more informative experiential knowledge.

Unlike static training on a fixed dataset, this iterative process enables the model to continuously refine its internalized knowledge by bootstrapping from its own improving behavior, naturally forming an online learning loop.
Importantly, each iteration only requires the model to interact with the user-side environment to collect new trajectories, while all training remains on the server side, making the process practical and scalable.
Algorithm~\ref{alg:oel} presents the pseudocode for the full iterative procedure.

\begin{algorithm}[t]
\small
\caption{Online Experiential Learning}
\label{alg:oel}
\begin{algorithmic}
\Require User-side environment $\mathcal{E}$; Model $\pi_\theta$
\Ensure Trained model $\pi_\theta$
\State \textcolor{purple}{\textit{Below presents the \textbf{Cross-Trajectory} variant. A simpler alternative \textbf{Self-Trajectory} variant, where knowledge extracted from each trajectory is applied to its own prefixes, is described in Algorithm~\ref{alg:oel_self}.}}
\While{Online Learning}
    \State \textcolor{blue}{[User Side]}
    \State Collect trajectories $\mathcal{T} = \{\tau_1, \ldots, \tau_n\}$ and $\mathcal{T}' = \{\tau_1, \ldots, \tau_m\}$ from $\mathcal{E}$ using $\pi_\theta$
    \Statex
    \State \textcolor{blue}{[Server Side]}
    \State \textcolor{gray}{\textit{// Stage 1: Extract Experiential Knowledge from User Trajectories}}
    \State Set $\pi_{\mathrm{extract}} = \pi_\theta$
    \State Accumulate experiential knowledge $\mathcal{C} = \{e^1, \ldots, e^K\}$ on $\mathcal{T}$ using $\pi_{\mathrm{extract}}$ via \Eqref{eq:extraction}
    \Statex
    \State \textcolor{gray}{\textit{// Stage 2: Consolidate Experiential Knowledge into Model Weights}}
    \State Construct partial rollout prefixes $\mathcal{D} = \{x_i^j\}$ from $\mathcal{T}'$
    \State Set $\pi_{\mathrm{teacher}} = \pi_\theta$ and keep it frozen 
    \For{batch $x \sim \mathcal{D}, e \sim \mathcal{C}$}
        \State Sample response $y \sim \pi_\theta(\cdot \mid x)$
        \State $\mathcal{L}(\theta) \gets \frac{1}{|y|} \sum_{t=1}^{|y|} { D_{\mathrm{KL}}\!\left( \pi_\theta(\cdot \mid x, y_{<t}) \,\Big\|\, \pi_{\mathrm{teacher}}(\cdot \mid e, x, y_{<t}) \right) } $
        \State Update $\theta$ by minimizing $\mathcal{L}(\theta)$ according to \Eqref{eq:objective}
    \EndFor
    \Statex
    \State Transfer updated $\pi_\theta$ to user side
\EndWhile
\State \Return $\pi_\theta$
\end{algorithmic}
\end{algorithm}

\section{Experiments}

\subsection{Setup}
\paragraph{Datasets and Models}
We conduct experiments on two text-based game environments, Frozen Lake and Sokoban, both implemented within TextArena~\citep{textarena}. In Frozen Lake, the agent navigates a grid to reach a goal location while avoiding holes. Sokoban is a spatial reasoning puzzle requiring the model to push a box onto a target position without falling into holes or getting stuck against walls.
No explicit rules are provided by the game; instead, the model must discover them through exploration~\citep{tencentgame,opcd}.
At each turn, TextArena returns a textual description of the resulting game state, such as whether a move was legal, hit a wall, led to a hole, or reached the goal, along with the updated map. This allows the language model to interact with the environment across multiple turns. Further details on the dataset are provided in Appendix~\ref{app:exp_detail_dataset}.
We use thinking models Qwen3-1.7B, Qwen3-4B, and Qwen3-8B~\citep{qwen3}, as well as a non-thinking model Qwen3-4B-Instruct-2507, to interact with the game environment.



\paragraph{Extraction Stage}
We set the extraction model to the deployed model of the current round, i.e., $\pi_{\mathrm{extract}} = \pi_\theta$. If the extraction model is a thinking model, thinking mode is enabled, and we retain the answer part as experiential knowledge while removing the reasoning part. We consider two formats of experiential knowledge: structured and unstructured. For the structured format, we prompt the extraction model $\pi_{\mathrm{extract}}$ to summarize transferable knowledge as a list of items, each prefixed with ``\texttt{-- EXPERIENCE ITEM:}'', retaining only entries that conform to this format. We set the number of trajectories for accumulation to $n=25$ or $n=50$, and the maximum generation length of the extractor to $L_{\max}=8192$ tokens.
For the unstructured format, the extractor generates knowledge freely without formatting constraints, with $n=15$ and $L_{\max}=2048$.
In both cases, $L_{\max}$ also serves as the maximum length of the resulting experiential knowledge; accumulated content exceeding this limit is truncated.
We repeat the accumulation process for $K=10$ times with different random seeds for both formats, resulting in a set of accumulated experiential knowledge $\mathcal{C}$.

Since the extraction process is performed server-side and we do not require scalar reward signals from the environment, we do not select the optimal experiential knowledge and instead retrieve the knowledge at the fixed accumulation step across \ours{} rounds.
Prompt templates are provided in Appendix~\ref{app:exp_detail_templates} and detailed configurations are provided in Appendix~\ref{app:exp_detail_extract}.

\paragraph{Consolidation Stage}
We perform on-policy context distillation for 20 or 100 steps per \ours{} round with 64 game samples per step, requiring 1280 or 6400 trajectory samples per training round. Each model interaction with the game environment spans up to 5 turns with a maximum response length of 1024 tokens per turn. For each training prefix, experiential knowledge $e$ is randomly sampled from $\mathcal{C}$.
We fix the number of training steps across all \ours{} rounds and adopt the final-step checkpoint without any checkpoint selection. We evaluate model performance using the pass rate on a held-out test split of size-128 game maps, averaged over 10 random seeds. For out-of-distribution evaluation, we report prompt-level strict accuracy on IF-Eval~\citep{ifeval}.
Further training details are provided in Appendix~\ref{app:exp_detail_train}.

\subsection{\ours{} Enables Online Learning}

\begin{figure}[t]
\centering
\includegraphics[width=0.99\linewidth]{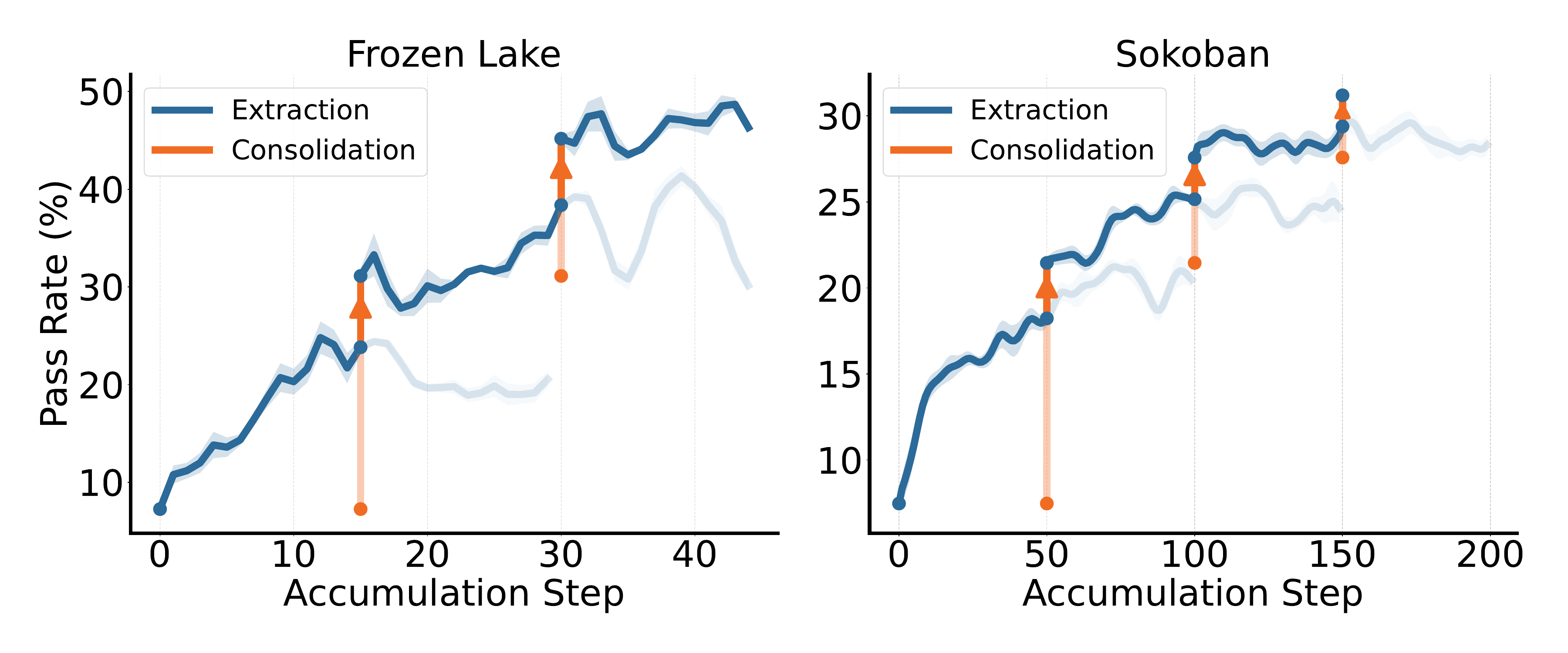}
\vspace{-0.1cm}
\caption{By iterating over experiential knowledge extraction and consolidation stages of \ours{}, the model can progressively improve pass rate, achieving online learning.}
\vspace{-0.1cm}
\label{fig:oel_curve_task}
\end{figure}

As shown in \Cref{fig:oel_curve_task}, by iterating over the experiential knowledge extraction and consolidation stages, \ours{} enables the model to progressively improve task performance on the online environment, effectively achieving online learning. We demonstrate this on Frozen Lake with a thinking model Qwen3-1.7B and on Sokoban with a non-thinking model Qwen3-4B-Instruct-2507.

During the accumulation phase, the pass rate steadily improves as experiential knowledge grows, but eventually saturates (transparent curves). This saturation is expected: as the experiential knowledge accumulates, the context window becomes increasingly occupied, limiting the model's capacity to absorb and leverage additional knowledge through in-context learning alone. Applying on-policy context distillation to consolidate at these intermediate points not only internalizes the accumulated experiential knowledge into model weights, but also surpasses the pre-consolidation performance. This is because the teacher model augmented with experiential knowledge serves as an effective reward model, providing dense token-level training signal that enables the student model to learn from consolidation training data that the teacher itself never accessed. 
In other words, the student can generalize beyond the teacher's in-context capabilities by distilling the knowledge directly into its parameters.

The consolidated model is then deployed for the next iteration, where its improved policy collects higher-quality trajectories. These trajectories contain richer information about successful strategies and failure modes, further boosting performance during subsequent accumulation. Notably, each new iteration starts from a stronger baseline, allowing the model to explore more challenging regions of the task space and extract increasingly sophisticated experiential knowledge. Across both settings, successive iterations of \ours{} yield consistent gains, demonstrating that the loop provides a robust mechanism for online learning without relying on any reward model or verifiable reward.

\subsection{\ours{} Improves Token Efficiency}

\begin{wrapfigure}{r}{7.0cm}
\centering
\vspace{-0.7cm}
\includegraphics[width=0.5\textwidth]{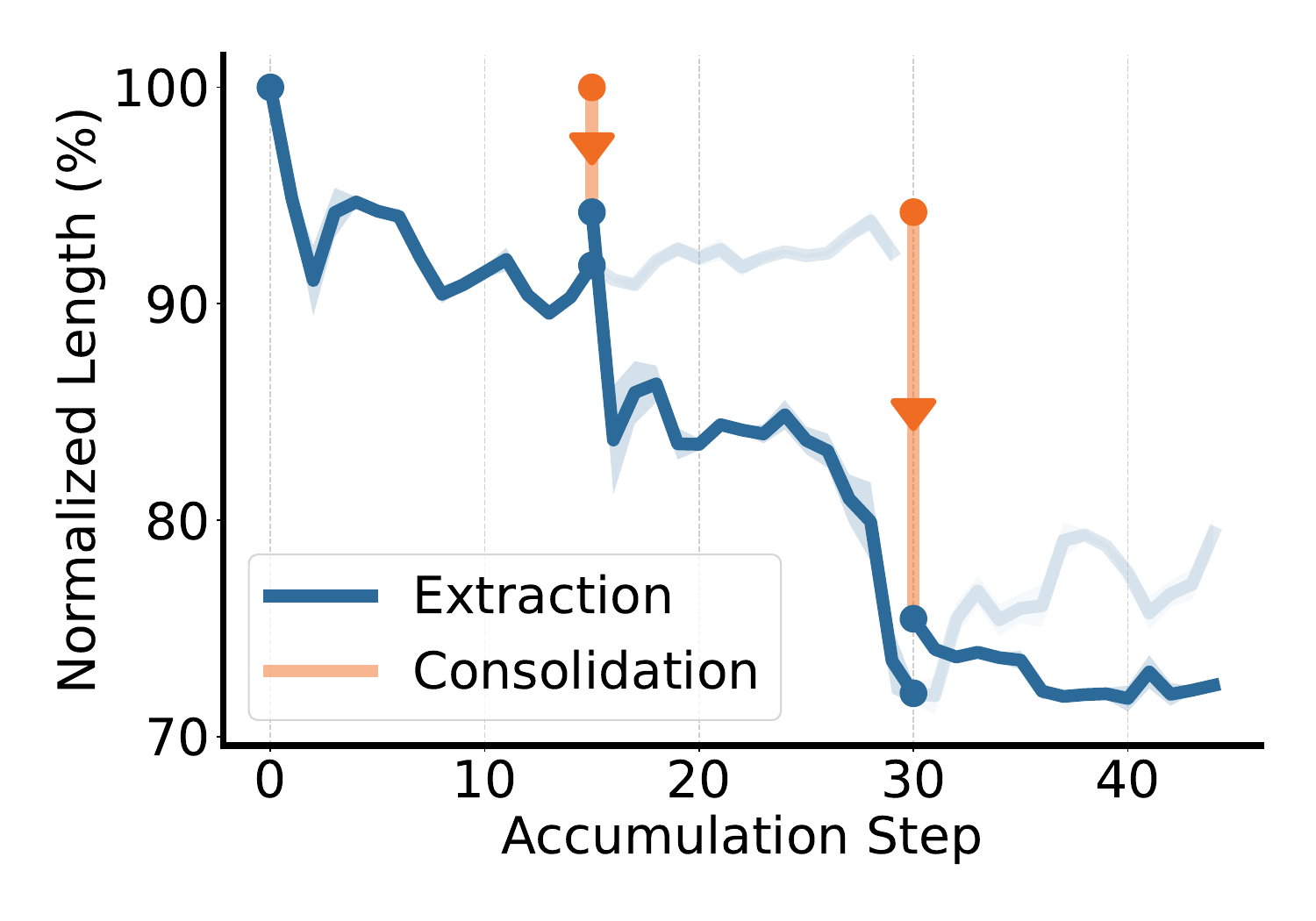}
\vspace{-0.45cm}
\caption{Normalized response length across \ours{} rounds. Reasoning becomes more efficient as experiential knowledge is progressively internalized.}
\vspace{-0.1cm}
\label{fig:oel_scaling_length}
\end{wrapfigure}

Beyond improving task performance, \ours{} also enables the model to solve problems faster over successive rounds. As shown in \Cref{fig:oel_scaling_length}, the average per-turn response length of Qwen3-1.7B on Frozen Lake decreases across accumulation steps, reducing to roughly 70\% of the initial length by the third iteration. During each extraction phase, the accumulated experiential knowledge helps the model arrive at correct answers faster. After consolidation, this pattern is retained in the model weights.
Combined with the concurrent pass rate improvements in \Cref{fig:oel_curve_task}, this confirms that successive iterations of \ours{} progressively internalize experiential knowledge, enabling the model to solve problems both more accurately and with less reasoning effort.

\subsection{\ours{} Mitigates Catastrophic Forgetting}

The on-policy context distillation used in \ours{} achieves better in-distribution performance while mitigating catastrophic forgetting on out-of-distribution tasks compared to off-policy context distillation. \ours{} employs on-policy context distillation during the consolidation stage, where training samples are generated from the policy model's own distribution.
In contrast, off-policy context distillation~\citep{context:distill:anthropic,context:distill:berkeley,infiniteicl} uses the teacher model equipped with experiential knowledge in context to generate responses, then minimizes the forward KL divergence between the context-free student model and the context-conditioned teacher on these collected responses to train the student. Since the responses are sampled from the knowledge-augmented model rather than the student itself, this constitutes off-policy training.

\begin{figure}
\centering
\includegraphics[width=\linewidth]{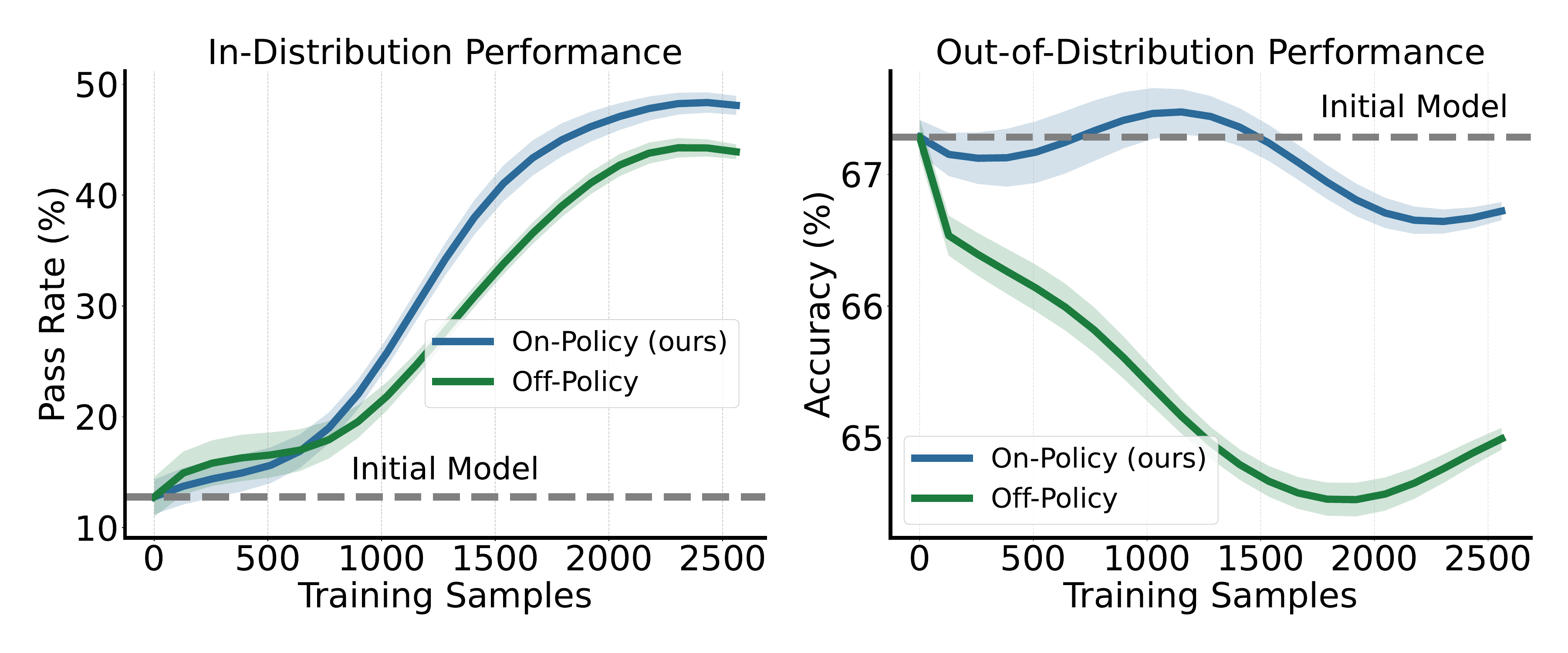}
\vspace{-0.15cm}
\caption{On-policy context distillation in \ours{} consolidation stage can achieve higher in-distribution (game pass rate) performance while better preserving out-of-distribution (IF-Eval accuracy) performance compared to off-policy context distillation.}
\vspace{-0.1cm}
\label{fig:oel_ood}
\end{figure}

We compare these two approaches in \Cref{fig:oel_ood}, using Qwen3-1.7B on FrozenLake. We concatenate Round~1 and Round~2 consolidation stages of 20 gradient steps and 64 batch size each; in-distribution performance tends to saturate within each stage after 20 steps, and we omit the saturated portions for clarity, applying smoothing to the concatenated curve.
The left subfigure shows in-distribution pass rate, while the right subfigure reports out-of-distribution (OOD) performance on IF-Eval. As shown, \ours{} achieves higher in-distribution performance than off-policy context distillation throughout training. More importantly, \ours{} largely preserves OOD performance close to the initial model, whereas off-policy context distillation exhibits a clear degradation over training steps. This is consistent with prior work showing that on-policy training mitigates catastrophic forgetting~\citep{rlrazor,retaining,opcd}, and confirms that the on-policy consolidation in \ours{} effectively internalizes experiential knowledge without sacrificing general capabilities.

\subsection{Effect of Model Size}

\begin{wrapfigure}{r}{7.1cm}
\centering
\vspace{-0.45cm}
\includegraphics[width=0.5\textwidth]{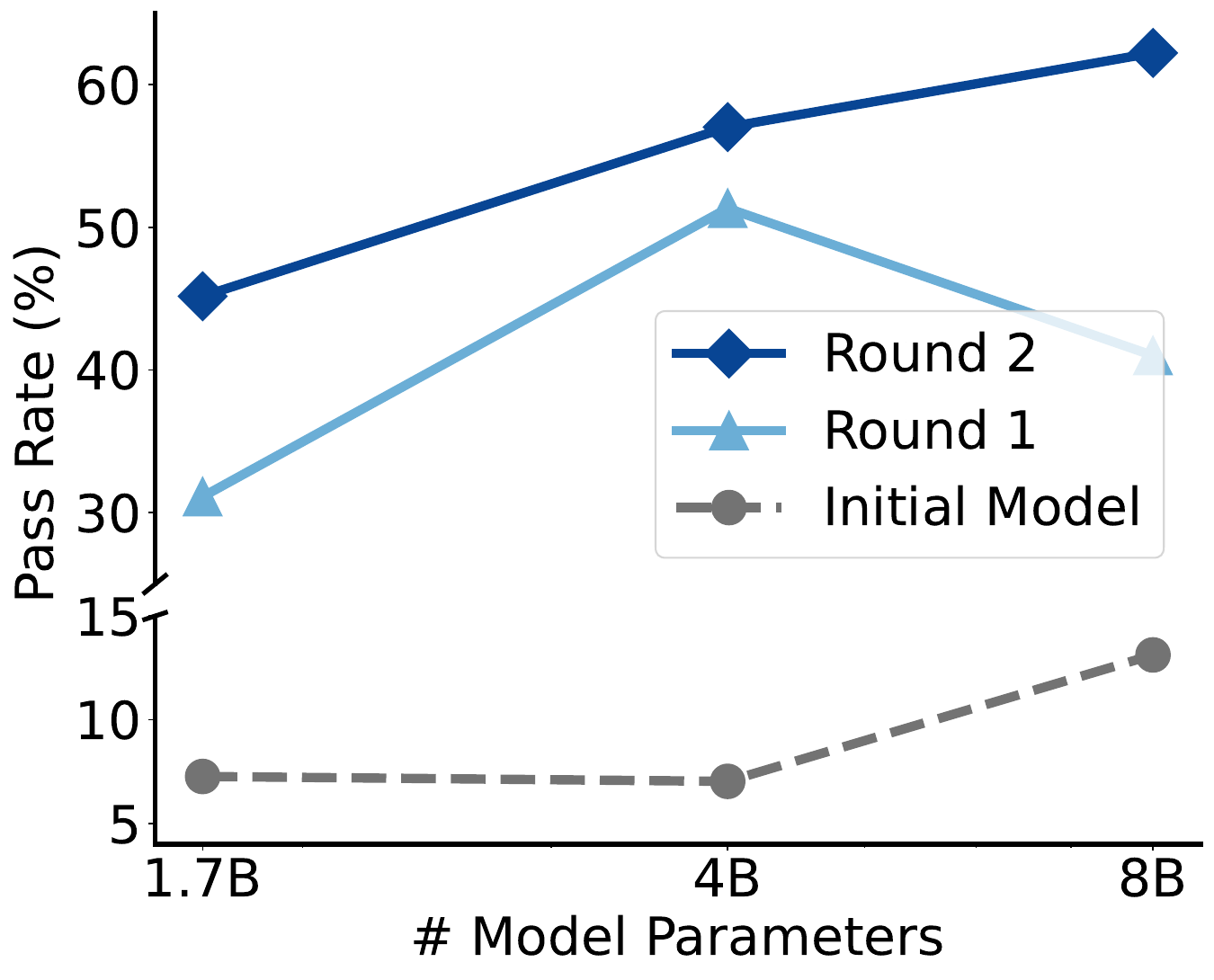}
\vspace{-0.45cm}
\caption{Performance scaling with model size across \ours{} rounds on Qwen3 models.}
\vspace{-0.1cm}
\label{fig:oel_scaling_size}
\end{wrapfigure}

We examine effect of model size of \ours{} in \Cref{fig:oel_scaling_size}, reporting the pass rate of Qwen3-1.7B, 4B, and 8B on FrozenLake across two rounds. While initial model performance remains relatively flat across scales, \ours{} yields substantial improvements for all model sizes, with larger models generally achieving higher pass rates.
Notably, the gain from Round~1 to Round~2 is consistent across scales, demonstrating that experiential knowledge continues to accumulate meaningfully beyond the first round regardless of model capacity.
Larger models generate higher-quality trajectories from which more effective experiential knowledge can be extracted, creating a virtuous cycle where greater capacity and better experience compound to amplify performance gains.

\subsection{Analysis}
\subsubsection{Learning from Experiential Knowledge over Raw Experience}

\begin{wrapfigure}{r}{7.0cm}
\centering
\begin{tabular}{@{}lcc@{}}
\toprule
 & \multicolumn{2}{c}{\textbf{Pass Rate (\%)}} \\
\cmidrule(l){2-3}
\textbf{Experience Type} & \textbf{In-Context} & \textbf{Consolidate} \\
\midrule
w/o Experience & \multicolumn{2}{c}{7.5} \\
\midrule
Raw Trajectory  & 10.9 & 7.8 \\
\textbf{Knowledge}  & \textbf{18.2} & \textbf{21.4} \\
\bottomrule
\end{tabular}
\makeatletter\def\@captype{table}\makeatother\caption{Extracted experiential knowledge context is more effective than raw trajectories for improving performance. Evaluated with Qwen3-4B-Instruct-2507 on Sokoban.}
\label{tab:game-rawexp}
\vspace{-1em}
\end{wrapfigure}



We validate the necessity of extracting experiential knowledge rather than directly using raw experience in \Cref{tab:game-rawexp}, reporting the pass rate of Qwen3-4B-Instruct-2507 on Sokoban in the first round. ``In-Context'' refers to prepending the experience to the context.
Simply using raw interaction trajectories yields only modest improvement, suggesting that unprocessed trajectories introduce noise that obscures useful information. In contrast, extracted experiential knowledge substantially improves the pass rate before and after consolidation, confirming that the extraction stage is essential for \ours{}.

\subsubsection{On-Policy Consistency Between Experiential Knowledge and Policy Model}

\begin{wrapfigure}{r}{7.0cm}
\centering
\begin{tabular}{@{}lcc@{}}
\toprule
 & \multicolumn{2}{c}{\textbf{Pass Rate (\%)}} \\
\cmidrule(l){2-3}
\textbf{Experience Source} & \textbf{In-Context} & \textbf{Consolidate} \\
\midrule
w/o Experience & \multicolumn{2}{c}{7.3} \\
\midrule
Qwen3-4B  & 18.0 & 22.7 \\
Qwen3-1.7B (\textbf{Self}) & \textbf{23.8} & \textbf{31.1} \\
\bottomrule
\end{tabular}
\makeatletter\def\@captype{table}\makeatother\caption{Performance of Qwen3-1.7B on Frozen Lake. On-policy experiential knowledge derived from its own trajectories benefits more than off-policy knowledge from a larger model Qwen3-4B.}
\label{tab:game-align}
\vspace{-1em}
\end{wrapfigure}



We investigate the importance of on-policy consistency between experiential knowledge and policy model in \Cref{tab:game-align}.
``In-Context'' refers to prepending the extracted experiential knowledge to the context.
Compared to experiential knowledge extracted from the larger Qwen3-4B, knowledge derived from Qwen3-1.7B's own trajectories yields higher pass rates.
This suggests that experiential knowledge from a stronger model does not necessarily transfer well, as it may encode strategies beyond the smaller model's capabilities. This highlights that on-policy consistency between experiential knowledge and the policy model is critical.


\section{Related Work}

\paragraph{On-Policy Distillation}
On-policy distillation methods~\citep{minillm,thinkingmachine-onpolicy,googlepolicy} train student models on their own generated trajectories rather than on teacher-produced data, mitigating the train-inference mismatch inherent in off-policy approaches. Minimizing the reverse KL divergence encourages mode-seeking behavior~\citep{minillm}. In \ours{}, on-policy distillation serves as the consolidation mechanism that internalizes accumulated experiential knowledge into model weights, with the additional benefit of preserving out-of-distribution performance compared to off-policy alternatives.

\paragraph{Context Distillation}
Context distillation aims to compress in-context knowledge into model parameters, removing the need to provide lengthy contexts at inference time~\citep{context:distill:anthropic,context:distill:berkeley,infiniteicl}. Typical approaches train a student model to imitate the outputs of a context-conditioned teacher using forward KL divergence on teacher-generated data. While effective for simple contexts, these off-policy methods can suffer from mode-covering behavior, particularly when the student lacks the capacity to fully capture the teacher's context-aware distribution.
\ours{} builds on on-policy context distillation~\citep{opcd}, which addresses these limitations by training on student-generated trajectories with reverse KL divergence.

\paragraph{Learning from Experience}
Learning from experience has long been a central theme in artificial intelligence. A recent position paper argues that agents should primarily learn from their own interaction with the world rather than from human-curated data, heralding an era of experience \citep{eraexp}. Along this direction, early-stage interaction experience has been shown to accelerate agent learning in subsequent tasks~\citep{earlyexp}, and reasoning-based agents have demonstrated the ability to discover game strategies through self-play and reflection~\citep{tencentgame}. In the language model community, several methods have explored leveraging interaction histories: ~\citep{reflexion} prompts models to reflect on past failures to guide future attempts, while~\citep{expel} extracts insights from trajectories and stores them in external memory for retrieval.

\section{Conclusion}

In this work, we introduced Online Experiential Learning (\ours{}), a reward-free framework that enables language models to continuously improve from their own deployment experience. By extracting transferable experiential knowledge from interaction trajectories and consolidating it into model parameters via on-policy context distillation, \ours{} forms a natural online learning loop that requires no human annotations, no reward models, and no server-side access to user environments. Our experiments on text-based game environments demonstrate that \ours{} achieves consistent improvements over successive iterations across multiple model scales and both thinking and non-thinking variants, enhancing task accuracy and inference efficiency while preserving out-of-distribution performance. Our analysis further confirms the importance of knowledge extraction over raw trajectories and the critical role of on-policy consistency. We believe online experiential learning represents a promising direction for the next stage of language model development, where real-world deployment serves not as the endpoint of training but as the beginning of continuous improvement.

\section*{Acknowledgements}

We are grateful to Yu Li and Yuxian Gu for discussions.

\bibliographystyle{alpha}
\bibliography{opcd}

\newpage
\appendix

\section{Implementation of On-Policy Context Distillation}
\label{app:opcd_detail}

Below we describe the formulation of on-policy context distillation~\citep{opcd} in detail.
Consider an input $x$ and a guiding context $c$ that is prepended to the input. The goal is to train a student model $\pi_\theta(\cdot \mid x)$ to match the behavior of a teacher model $\pi_\mathrm{teacher}(\cdot \mid c, x)$ that has access to $c$.
Specifically, the training objective minimizes the reverse Kullback-Leibler (KL) divergence between the two distributions, where responses are sampled on-policy from the student.
By decomposing the sequence-level divergence into a sum over individual token positions, we obtain the following loss:
\begin{equation}
\mathcal{L}(\theta) = \mathbb{E}_{(x, c) \sim \mathcal{D}, y \sim \pi_\theta(\cdot \mid x)} \left[ \frac{1}{|y|} \sum_{t=1}^{|y|} { D_\mathrm{KL} \left( \pi_\theta(\cdot \mid x, y_{<t}) \| \pi_\mathrm{teacher}(\cdot \mid c, x, y_{<t}) \right) } \right]
\label{eq:opcd-objective}
\end{equation}
Here, $c$ represents the in-context knowledge that the student aims to internalize, $\mathcal{D}$ denotes the training dataset, and $y$ is a response sampled from the current student policy.

At each token position, the reverse KL divergence is computed as:
\begin{equation}
\begin{aligned}
&D_\mathrm{KL} \left( \pi_\theta(\cdot \mid x, y_{<t}) \| \pi_\mathrm{teacher}(\cdot \mid c, x, y_{<t}) \right) \\
=&\ \mathbb{E}_{y_t' \sim \pi_\theta(\cdot \mid x, y_{<t})} \left[    \log \frac{\pi_\theta(y_t' \mid x, y_{<t})}{\pi_\mathrm{teacher}(y_t' \mid c, x, y_{<t})} \right] \\
=& \sum_{y_t' \in \mathcal{V}} \pi_\theta(y_t' \mid x, y_{<t} ) { \left( \log \pi_\theta(y_t' \mid x, y_{<t}) - \log  \pi_\mathrm{teacher}(y_t' \mid c, x, y_{<t}) \right) }
\label{eq:token:kl}
\end{aligned}
\end{equation}
where $\mathcal{V}$ is the vocabulary.
For computational efficiency, we approximate the full summation over $\mathcal{V}$ by considering only the top-$k$ tokens ranked by the student's predicted probability, denoted $\mathcal{V}_{\operatorname{top-k}}$. Throughout all experiments, we set $k=256$.

\newpage
\section{Self-Trajectory Variant}
\label{app:self_trajectory}

In the \textbf{Cross-Trajectory} variant (used in our experiments) presented in Algorithm~\ref{alg:oel}, experiential knowledge is accumulated across a set of trajectories $\mathcal{T}$ and then applied to partial rollout prefixes constructed from a separate set $\mathcal{T}'$. This decouples knowledge extraction from training data construction, allowing the extracted knowledge to generalize across different trajectories.

We also consider a simpler alternative \textbf{Self-Trajectory} variant, where only a single set of trajectories $\mathcal{T}$ is collected. In this variant, experiential knowledge is extracted independently from each trajectory $\tau_i$, and the resulting knowledge $e_i$ is paired exclusively with the partial rollout prefixes derived from the same trajectory. This creates a tighter coupling between the extracted knowledge and the training context, as the knowledge directly reflects the experience from which the prefixes originate. The full procedure is given in Algorithm~\ref{alg:oel_self}.

\begin{algorithm}[h]
\small
\caption{Online Experiential Learning (Self-Trajectory Variant)}
\label{alg:oel_self}
\begin{algorithmic}
\Require User-side environment $\mathcal{E}$; Model $\pi_\theta$
\Ensure Trained model $\pi_\theta$
\Statex
\While{Online Learning}
    \State \textcolor{blue}{[User Side]}
    \State Collect trajectories $\mathcal{T} = \{\tau_1, \ldots, \tau_n\}$ from $\mathcal{E}$ using $\pi_\theta$
    \Statex
    \State \textcolor{blue}{[Server Side]}
    \State \textcolor{gray}{\textit{// Stage 1: Extract Experiential Knowledge from User Trajectories}}
    \State Set $\pi_{\mathrm{extract}} = \pi_\theta$
    \For{each $\tau_i \in \mathcal{T}$}
        \State Extract experiential knowledge $e_i \sim \pi_{\mathrm{extract}}(\cdot \mid \tau_i)$
    \EndFor
    \Statex
    \State \textcolor{gray}{\textit{// Stage 2: Consolidate Experiential Knowledge into Model Weights}}
    \State Construct partial rollout prefixes $\mathcal{D} = \{(x_i^j,\, e_i)\}$ from $\mathcal{T}$
    \State Set $\pi_{\mathrm{teacher}} = \pi_\theta$ and keep it frozen
    \For{batch $(x,\, e) \sim \mathcal{D}$}
        \State Sample response $y \sim \pi_\theta(\cdot \mid x)$
        \State $\mathcal{L}(\theta) \gets \frac{1}{|y|} \sum_{t=1}^{|y|} { D_{\mathrm{KL}}\!\left( \pi_\theta(\cdot \mid x, y_{<t}) \,\Big\|\, \pi_{\mathrm{teacher}}(\cdot \mid e, x, y_{<t}) \right) } $
        \State Update $\theta$ by minimizing $\mathcal{L}(\theta)$ according to \Eqref{eq:objective}
    \EndFor
    \Statex
    \State Transfer updated $\pi_\theta$ to user side
\EndWhile
\State \Return $\pi_\theta$
\end{algorithmic}
\end{algorithm}

\newpage
\section{Details of Experiments}
\label{app:exp_detail}

\subsection{Dataset Details}
\label{app:exp_detail_dataset}
Frozen Lake and Sokoban are two text-based game environments built on TextArena~\citep{textarena}. In Frozen Lake, the agent navigates a grid to reach a goal location while avoiding holes; we use a 3 $\times$ 3 grid with two holes in our experiments.
Sokoban is a spatial reasoning puzzle that requires the model to push a box onto a target position without falling into holes or getting stuck against walls; we use a 6 $\times$ 6 grid with one box.

Neither game provides explicit rules, so the model must discover them through exploration. Following~\citep{tencentgame,opcd}, we replace the original rules provided by TextArena with a general task description, as illustrated in \Cref{fig:replace_rule_fz} and \Cref{fig:replace_rule_sok}. This setup simulates real-world scenarios where the model receives minimal prior knowledge about a new environment.

\begin{figure}[ht]
    \begin{tcolorbox}
    \textit{Replace:} \\
    Welcome to Frozen Lake!\textbackslash n\textbackslash nYou are represented by 'P' on the grid.\textbackslash nGrid symbols:\textbackslash n' ' = Frozen surface (safe to walk on)\textbackslash n'H' = Hole (fall in and lose!)\textbackslash n'G' = Goal (reach this to win!)\textbackslash n'P' = Your current position\textbackslash n\textbackslash nAvailable actions: up, down, left, right (or w, a, s, d)\textbackslash nType your action as: [up], [down], [left], [right] or [w], [a], [s], [d]\textbackslash n\textbackslash nObjective: Navigate from the start (top-left) to the goal (bottom-right) without falling into any holes!\textbackslash n\textbackslash n \\ \\
    \textit{With:} \\
    You are the player and you are represented by 'P' on the grid. You should select the best action to reach the goal in the shortest number of steps. Your only way to interact is to move one step each time. Available actions: up, down, left, right (or w, a, s, d). Type your action as: [up], [down], [left], [right] or [w], [a], [s], [d]\textbackslash n
    \end{tcolorbox}
    \caption{The game environment provides no explicit rules to simulate real-world scenarios where models receive minimal prior information about new environments. We process initial prompt of Frozen Lake from TextArena to replace explicit rules.}
    \label{fig:replace_rule_fz}
\end{figure}

\begin{figure}[ht]
    \begin{tcolorbox}
    \textit{Replace:} \\
    You are solving the Sokoban puzzle. You are the player and you need to push all boxes to targets.\textbackslash n~~~~~~~~When you are right next to a box, you can push it by moving in the same direction.\textbackslash n~~~~~~~~You cannot push a box through a wall, and you cannot pull a box.\textbackslash n~~~~~~~~On the board, objects are represented as: \textbackslash n~~~~~~~~- The player (you) appears as 'P' \textbackslash n~~~~~~~~- Walls are represented with '\#' \textbackslash n~~~~~~~~- Boxes are marked as 'X' \textbackslash n~~~~~~~~- Empty goals are shown with a 'O'\textbackslash n~~~~~~~~- Boxes on goals are visualized with '$\sqrt{}$'\textbackslash n~~~~~~~~You can also use [w] for up, [a] for left, [s] for down, and [d] for right. \\ \\
    \textit{With:} \\
    You are the player and you are represented by 'P' on the grid. You should select the best action to reach the goal in the shortest number of steps. Your only way to interact is to move one step each time. Available actions: up, down, left, right (or w, a, s, d). Type your action as: [up], [down], [left], [right] or [w], [a], [s], [d]\textbackslash n
    \end{tcolorbox}
    \caption{We process initial prompt of Sokoban from TextArena to replace explicit rules.}
    \label{fig:replace_rule_sok}
\end{figure}

At every turn, TextArena returns a textual observation describing the outcome of the action taken (e.g., whether it was legal, collided with a wall, led to a hole, or reached the goal) together with the updated map, allowing the language model to engage with the environment over multiple turns.

\newpage
\subsection{Prompt Templates}
\label{app:exp_detail_templates}
When converting experience into experiential knowledge, we consider two formats: structured and unstructured. For the structured format, we prompt the extraction model $\pi_{\mathrm{extract}}$ to summarize transferable knowledge as a list of items, each prefixed with ``\texttt{-- EXPERIENCE ITEM:}'', retaining only entries that conform to this format. For the unstructured format, the extractor generates knowledge freely without any formatting constraints.

For the structured format, we use the prompt template in \Cref{fig:exp_accum_game_structured}. The input context ``\texttt{latest\_experience}'' comprises the multi-turn game environment outputs, and the model responses (including the thinking process when available), and ``\texttt{previous\_experience}'' is previously accumulated experiential knowledge. We then extract lines prefixed with ``\texttt{-- EXPERIENCE ITEM:}'' as valid experiential knowledge items.

\begin{figure}[ht]
    \begin{tcolorbox}
    You are an AI language model that continuously refines its internal experience. \\
    Here is the interaction history (the game environment (input) and your response and action (output)): \\
    \{latest\_experience\} \\ \\
    Here is the previous experience: \\
    \# Experience \\
    \{previous\_experience\} \\ \\
    Your task: \\
    Based on the multi-round interaction history, generate experience for future learning. You should conduct a deep, comparative analysis to infer the game rules and the fundamental principles behind winning and losing. Using the interaction history and environment feedback, hypothesize the game rules and effective winning strategies, and organize these insights into 1-2 concise, high-level, and widely applicable experience items that help the player succeed in the game. \\ \\
    Rules: \\
    - The experience you generate MUST be formatted strictly as a markdown item which starts with "- EXPERIENCE ITEM:": \\
    - EXPERIENCE ITEM: ... \\
    - EXPERIENCE ITEM: ... \\
    - The experience you generate will be directly appended to the previous experience. Do not repeat the previous experience. Make sure the newly generated experience is different from the previous experience. \\
    - Your generated experience should be possible rules, instructions or winning strategies for the game. The experience should be generally useful rather than only applicable for the current map (board). \\  \\
    After careful reasoning step by step, output the final result in exactly this format: \\ \\
    Additional Experience (Rules or Strategies): \\
    \# Experience \\
    - EXPERIENCE ITEM: ...
    \end{tcolorbox}
    \caption{The prompt wrapper for structured experiential knowledge extraction on text games.}
    \label{fig:exp_accum_game_structured}
\end{figure}

For the unstructured format, we use the prompt template in \Cref{fig:exp_accum_game_unstructured}.

\begin{figure}[ht]
    \begin{tcolorbox}
    You are an AI language model that continuously refines its internal experience. \\
    Here is the interaction history (previous experience, the game environment (input) and your response and action (output)): \\
    \{latest\_experience\} \\ \\
    Here is the previous experience: \\
    \# Experience \\
    \{previous\_experience\} \\ \\
    Your task: \\
    Based on the multi-round interaction history and the previous experience, generate experience for future learning. You should conduct a deep, comparative analysis to infer the game rules and the fundamental principles behind winning and losing. Using the interaction history and environment feedback, hypothesize the game rules and effective winning strategies, and organize these insights into experience items that help the player succeed in the game. \\ \\
    Rules: \\
    - The experience you generate will be directly appended to the previous experience. Do not repeat the previous experience. Make sure the newly generated experience is different from the previous experience. \\
    - Your generated experience should be possible rules, instructions or winning strategies for the game. The experience should be generally useful rather than only applicable for the current map (board). \\  \\
    After careful reasoning step by step, output the final additional experience.
    \end{tcolorbox}
    \caption{The unstructured prompt wrapper for experiential knowledge extraction on text games.}
    \label{fig:exp_accum_game_unstructured}
\end{figure}

For new problems we embed experiential knowledge with the prompt template in \Cref{fig:exp_solve}.

\begin{figure}[ht]
    \begin{tcolorbox}
    You are an agent playing a game on a grid, acting as a reasoning engine.\\ \\
    Your decisions are based on the experience you have learned about the game's rules or strategies. This experience is only a guess of how the game works, and the rules and strategies may be incomplete or incorrect. \\ \\
    Given experience for rules or strategies you have learned: \\
    \{experience\} \\ \\
    Current situation: \\
    \{prompt\} \\ \\
    What action do you take? (Remember to wrap your final answer of the action in square brackets)
    \end{tcolorbox}
    \caption{The prompt wrapper for new problem solving with accumulated experiential knowledge.}
    \label{fig:exp_solve}
\end{figure}

\newpage
\subsection{Extraction Stage}
\label{app:exp_detail_extract}
For the structured format, we prompt the extraction model $\pi_{\mathrm{extract}}$ to summarize transferable knowledge as a list of items, each prefixed with ``\texttt{-- EXPERIENCE ITEM:}'', retaining only entries that conform to this format. We set the number of trajectories for accumulation to $n=25$ or $n=50$, and the maximum generation length of the extractor to $L_{\max}=8192$ tokens.
For the unstructured format, the extractor generates knowledge freely without formatting constraints, with $n=15$ and $L_{\max}=2048$.
In both cases, $L_{\max}$ also serves as the maximum length of the resulting experiential knowledge; accumulated content exceeding this limit is truncated.
We repeat the accumulation process for $K=10$ times with different random seeds for both formats, resulting in a set of accumulated experiential knowledge $\mathcal{C}$.
See \Cref{tab:hyperparam_search} and \Cref{tab:hyperparam_used} for more details.

Since the extraction process is performed server-side and we do not require scalar reward signals from the environment, we do not select the optimal experiential knowledge and instead retrieve the knowledge at the fixed accumulation step across \ours{} rounds.

\subsection{Consolidation Stage}
\label{app:exp_detail_train}

We perform on-policy context distillation for 20 or 100 steps per \ours{} round with 64 game samples per step, requiring 1280 or 6400 trajectory samples per training round. Each model interaction with the game environment spans up to 5 turns with a maximum response length of 1024 tokens per turn. For each training prefix, experiential knowledge $e$ is randomly sampled from $\mathcal{C}$.
We fix the number of training steps across all \ours{} rounds and adopt the final-step checkpoint without any checkpoint selection. We evaluate model performance using the pass rate on a held-out test split of size-128 game maps, averaged over 10 random seeds. For out-of-distribution evaluation, we report prompt-level strict accuracy on IF-Eval~\citep{ifeval}.

We compute the reverse KL divergence using the top 256 vocabulary tokens with the highest student model probabilities. We search learning rate in [1e-6, 5e-6] for different model and task configurations. The learning rate remains fixed across \ours{} rounds. The sampling temperature is set to 0.7.

\Cref{tab:hyperparam_search} presents the hyperparameter search ranges in extraction and consolidation stages. \Cref{tab:hyperparam_used} summarize the final configurations used for each model-task pair in our experiments.
All hyperparameters are fixed across \ours{} rounds of a single model-task pair.
For Qwen3-1.7B, we do not include previously accumulated experiential knowledge in the extraction context, as we find that smaller models lack sufficient capacity to effectively leverage long contextual information. For all other configurations, the extraction prompt includes experiential knowledge from previous accumulation steps.

\begin{table}[t]
\centering
\small
\begin{tabular}{@{}lc@{}}
\toprule
\textbf{Hyperparameter} & \textbf{Search Range} \\
\midrule
Knowledge Format & \{Structured, Unstructured\} \\
Structured $n$ & $\{25,\ 50\}$ \\
Unstructured $n$ & 15 \\
Structured $L_{\max}$ & 8192 \\
Unstructured $L_{\max}$ & 2048 \\
\midrule
Learning Rate & $\{1\mathrm{e}{-6},\ 5\mathrm{e}{-6}\}$ \\
Training Steps Each Round & $\{20,\ 100\}$ \\
\bottomrule
\end{tabular}
\vspace{0.2cm}
\caption{Hyperparameter search ranges for the extraction and consolidation stages.}
\label{tab:hyperparam_search}
\end{table}

\begin{table}[t]
\centering
\small
\begin{tabular}{@{}lcccc@{}}
\toprule
\bf Hyperparameter & \tabincell{c}{\textbf{Qwen3-1.7B} \\ \textbf{Frozen Lake}} & \tabincell{c}{\textbf{Qwen3-4B} \\ \textbf{Frozen Lake}} & \tabincell{c}{\textbf{Qwen3-8B} \\ \textbf{Frozen Lake}} & \tabincell{c}{\textbf{Qwen3-4B-Instruct} \\ \textbf{Sokoban}} \\
\midrule
Knowledge Format & Unstructured & Structured & Structured & Structured \\
Accumulated $n$ & 15 & 25 & 50 & 50 \\
Knowledge Length $L_{\max}$ & 2048 & 8192 & 8192 & 8192 \\
\midrule
Learning Rate & $5\mathrm{e}{-6}$ & $1\mathrm{e}{-6}$ & $1\mathrm{e}{-6}$ & $1\mathrm{e}{-6}$ \\
Training Steps & 20 & 20 & 100 & 100 \\
\bottomrule
\end{tabular}
\vspace{0.2cm}
\caption{Hyperparameters used for each model and task configuration. All values are fixed across \ours{} rounds for a single model-task pair.}
\label{tab:hyperparam_used}
\end{table}

\newpage
\section{Experiential Knowledge Examples}

We provide some experiential knowledge examples for Sokoban using Qwen3-4B-Instruct-2507 in \Cref{fig:exp_sok_example}.

\begin{figure}[ht]
    \begin{tcolorbox}
    - EXPERIENCE ITEM: The game's core mechanic is "axis-aligned convergence": at every move, the player must reduce the Manhattan distance to the goal by progressing directly along the shared row or column. This ensures that each action contributes to directional progress, prevents stagnation, and guarantees a shortest-path solution in static, obstacle-filled environments. Movement must always be toward the goal's row or column—any lateral or non-aligned step increases distance and risks entrapment in isolated zones. This principle transforms navigation from an exploratory process into a deterministic, rule-based sequence of progressive advancement toward the endpoint. \\

    - EXPERIENCE ITEM: A critical failure mode occurs when players rely on hypothetical or future board states during path planning—this leads to invalid moves (e.g., moving into walls or obstacles) that result in immediate termination. Success hinges on a reactive, state-dependent decision-making process: every move must be validated against the current board state before submission, with explicit attention to cell content (wall '\#', obstacle 'X', empty '\_' or goal 'O') and correct syntax (e.g., [up], [down]) to ensure validity and avoid game termination. This enforces a zero-tolerance policy for assumptions about future or hypothetical configurations, requiring players to inspect each move's destination before action. \\

    - EXPERIENCE ITEM: The game's true objective is not to reach any blank space ('\_'), but specifically to reach the goal marked by 'O'—this symbol serves as the sole valid endpoint. All path planning must be anchored to the position of 'O', and every move must be directed toward reducing the Manhattan distance to it. Any deviation—such as moving laterally when the goal is off-axis—increases distance and risks entrapment, especially in confined or symmetric layouts where obstacles force detours. This makes the identification and constant tracking of the 'O' position a foundational requirement for success. \\

    - EXPERIENCE ITEM: A universal principle of success is "progressive alignment through convergence": the player must continuously assess whether their current position shares a row or column with the goal ('O'). If aligned, move directly toward it; if not, move to the nearest adjacent cell that restores alignment—this ensures directional consistency, minimizes total steps, and prevents looping. This strategy transforms navigation from an exploratory process into a deterministic, rule-based sequence of efficient advancement, making it universally applicable across all board configurations and ensuring optimal path efficiency even when obstacles block direct paths.
    \end{tcolorbox}
    \caption{Some experiential knowledge examples for Sokoban game.}
    \label{fig:exp_sok_example}
\end{figure}

\end{document}